\def\year{2016}\relax
\documentclass[letterpaper]{article}
\usepackage{aaai16}
\usepackage{times}
\usepackage{helvet}
\usepackage{courier}
\usepackage{subcaption}
\usepackage{graphicx}
\begin{document}
% The file aaai.sty is the style file for AAAI Press 
% proceedings, working notes, and technical reports.
%
\title{PCG-Based Game Design Patterns}
\author{
Michael Cook\\
University of Falmouth\\
\And
Mirjam Eladhari\\
Otter Play Games\\
\And
Andy Nealen\\
NYU Game Center\\
\And
Mike Treanor\\
American University\\
\AND
Eddy Boxerman\\
Hemisphere Games\\
\And
Alex Jaffe\\
Spry Fox\\
\And
Paul Sottosanti\\
Riot Games\\
\And
Steve Swink\\
Cube Heart\\
}

\maketitle

\begin{abstract}
People enjoy encounters with generative software, but rarely are they encouraged to interact with, understand or engage with it. In this paper we define the term \textit{PCG-based game}, and explain how this concept follows on from the idea of an AI-based game. We look at existing examples of games which foreground their AI, put forward a methodology for designing PCG-based games, describe some example case study designs for PCG-based games, and describe lessons learned during this process of sketching and developing ideas.
\end{abstract}

\section{Introduction}

\noindent Generative software, whether in games or outside, is a source of delight and entertainment for users. The popularity of Twitter bots is a good example of how people are deriving enjoyment from viewing and experiencing the output of generative systems. This is particularly true of games, where generators have become increasingly common and people who regularly play games are more and more comfortable with the idea of `generated content' and what that might entail. Players of the game \textit{Minecraft} \cite{minecraft} collect the random seed integers that describe worlds of particular beauty and archive them online, while roguelike players repeatedly generate and discard worlds until they find one with particular features. We are familiar with generative systems and we enjoy controlling, interacting with and exploring them.

Despite this, games traditionally hide generative systems away from the player. Level generators are typically non-interactive providers of content, and where they are controllable it is generally through a menu prior to gameplay -- there is a clear delineation in the game's design between \textit{setting up} the generator before \textit{playing} the game proper. Generative systems are designed to passively fill out a game world with content, rather than being a focus of the player's time and attention, or even the purpose of playing the game in the first place. 

The term \textit{AI-based game} \cite{eladhari2011,treanor2015} was coined to describe games which \textit{foreground} an AI system of some kind. By this we mean that the game makes an AI system especially visible to the player, and the primary focus of the game is to interact with or be affected by the AI system in some way. Some examples of AI-based games include \textit{Alien: Isolation} \cite{alien} in which the player's relationship with a single AI enemy is the focus of gameplay, or \textit{Third Eye Crime} \cite{thirdeye} where pathfinding algorithms are visually represented to the player as the key skill that allows them to solve the game's puzzles.

In this paper we refine the notion of AI-based games into a special case of games driven by procedural generators. We call these \textit{PCG-based games}, and describe concretely how they embody the original premise of AI-based games. We describe why these are uniquely important among AI-based games, and why the opportunity for literacy and interaction with procedural generators make them valuable cultural artifacts and learning tools, as well as fertile ground for new game ideas. Finally, we describe some example game proposals we developed, and show how they highlight common issues or concepts with using procedural generation as the central focus of a game.

\section{Related Work}
Smith et al. describe \textit{Endless Web} \cite{endlessweb}, as a platforming game with procedurally generated levels. By exploring the levels and choosing particular exits, the player can alter the parameters to the procedural generator, allowing them to explore the generative space through gameplay. This is the earliest example we are aware of in which a game is designed around a procedural generator with the explicit intention of giving a player control of the generator's output. This is perhaps the best-known and most explicit use of procedural generation as a game mechanic. 

Many games expose a procedural generator of some kind to the player, usually in a preparatory phase before gameplay begins. \textit{Civilisation V} \cite{civ} is one such game, allowing the player to customise features of the world generation algorithm such as erosion strength, global temperature averages, sea levels, and resource distribution. This allows the player to roughly shape what content is generated, sometimes as a way of specialising the difficulty curve, and at other times as a form of abstract self-expression (deciding to play in a particular world, or on a planet with a custom backstory). 

The framing of player control over generators inside games is almost always explicit and literal - Minecraft refers to the act of `generat[ing] a new world' when its generative algorithm is invoked, while \textit{Invisible Inc.} offers players a `custom campaign settings' window where options can be altered before starting a campaign. This is important, as it distinguishes the work from our aims here by separating the interactions with a generator from the act of playing a game. In physical games, generator configuration is typically done prior to gameplay too -- deck construction in games such as \textit{Netrunner} \cite{netrunner} or \textit{Magic: The Gathering} \cite{magic} are ways of configuring the space of shuffled decks from which the player will later draw from, and games like \textit{Carcassonne} or \textit{Dungeon Run} allow the players to add or remove types of card from a deck to change the resulting play spaces created by the game. As with digital games, these are all setup activities (although deck-building and drafting is arguably a fundamental part of gameplay \cite{drafting}).

\begin{figure}
\begin{center}
\includegraphics[width=0.9\columnwidth]{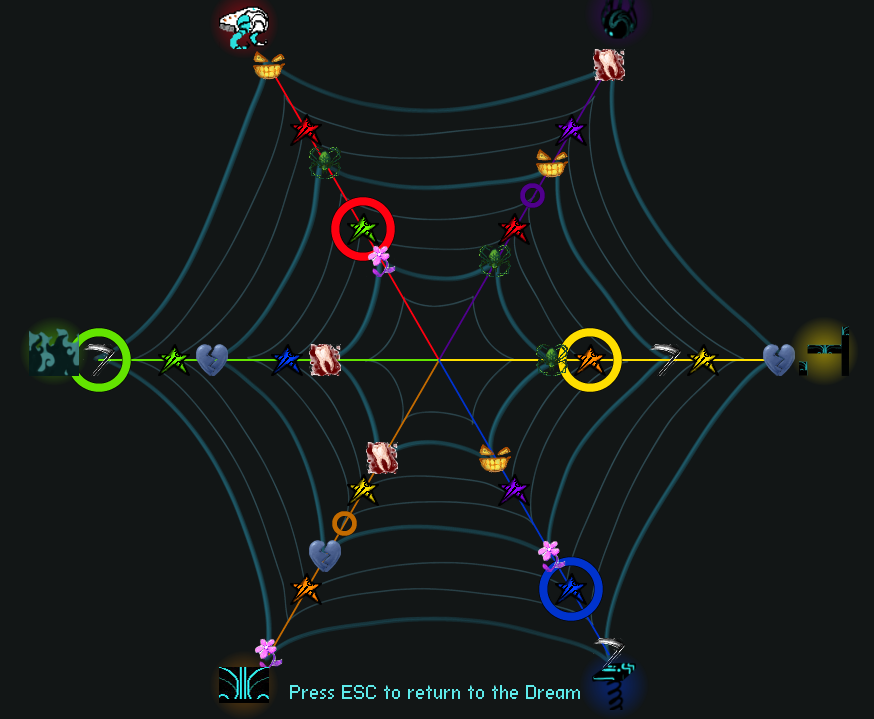}
\end{center}
\caption{A screenshot showing part of Endless Web's in-game `map'. Each line extending outwards from the center of the web represents a vector along which the level generator can be modified.}
\label{ewshot}
\end{figure}

Many researchers have looked at the issues that arise when people interact with procedural generators. In \cite{khaled} Khaled explores the metaphors designers use when talking about procedural generation. Even among only professional designers this is quite diverse, and includes metaphors such as \textit{tool}, \textit{material} and \textit{designer}. Given the breadth of uses for generative techniques, and the varying levels of complexity to which they are employed, we imagine there are many more concepts about generative systems held by people who play and interact with them. Part of this work's aims is to develop games that bring these issues to the fore, and allow us to study user understanding.

Elsewhere, work in \cite{shaker2010towards} or in \cite{mumford-iccc2015} show the relationship between users and generative systems, especially in the context of games or game-like applications. \cite{mumford-iccc2015} sheds light on, for example, how the presentation of generated content affects our perception of the system which generated it. Games like \textit{Dwarf Fortress} allow us to watch the slow creation process of a world, which may makes us feel differently about the quality and intelligence of the system than if it had appeared out of nowhere. 

\section{A Design Taxonomy For PCG}
In this section we deconstruct the notion of generative techniques for games into a taxonomy that is based on the qualities we might be interested in as prospective game designers. The best-known taxonomy for procedural generation is perhaps \cite{togelius}, which categorises approaches based on dimensions relating to when they are used, what kind of content they create, and also how controllable they are. Here we focus instead on the affordances and nature of the generator from the perspective of a game designer.

\subsection{An Interaction Taxonomy}
Different generators have different means by which they can be edited and meaningfully changed. This taxonomy outlines a non-exhaustive list of ways in which common types of generator can be altered either prior to, during, or after execution. 

\subsubsection{Starting State} Some generators, such as cellular automata systems or L-Systems, iterate upon a starting state of some kind in order to generate their content, typically expressed in the same format as its final output. For these systems, the output content can be dramatically changed and controlled simply by changing this starting state, even if all parameters and random seeds are left unchanged. 

\subsubsection{Parameters} Many generators have controllable parameters in the form of boolean fields, numerical ranges, or a random seed. Changing these parameters can have a range of effects on the output. The number of generations in an evolutionary system, or the size of the array in the Diamond-Square algorithm, are simple examples of generator parameters.

\subsubsection{Rules} While most generators embody some notion of a procedure for generating content, certain algorithms represent an abstract set of rules that define a crucial part of their generative process. L-Systems and context-free design grammars are two examples of systems which have internal representations of rules which guide their generation.

\subsubsection{Objective Functions} Generate-and-test approaches evaluate generated content and either discard or repurpose content which does not meet the standards it has. This process is normally separate to the act of generation (although in some cases it is woven in, as with an evolutionary system's fitness function) but altering it affects the kind of content the generator produces.

\subsection{A Content Taxonomy}
Generators are used for an increasingly wide range of purposes within games, expanding beyond classical uses of environment creation and item randomisation. Here we (non-exhaustively) taxonomise procedural generators according to the role that the generated content plays within a game design. The intent here is to help classify approximately what aspect of the game's systems the player will be interacting with and changing.

\subsubsection{Progression Systems}
Generators which produce content tied into an escalation of difficulty and reward are influencing a game's progression systems. A generator which creates items found on an adventure is tied into the player's gradual increase in strength - stronger and more plentiful items result in a more powerful player or one with more resources. Other examples include games which generate enemies according to approximate difficulty curves.

\subsubsection{Environment \& Space}
Generators which produce levels, worlds or other explorable spaces. This might be specific-case generators such as a roguelike's dungeon generation algorithm where the player path is often tightly incorporated into the design, or it might be a more open environment generator such as Minecraft's world generator. This content often sets specific challenges for the player, either in traversal (mastering abilities like jumping or navigation) or exploration (finding a particular item, place or resource in a large area). 

\subsubsection{Narrative}
Generators which either produce sequences of events framed as a story, or those which simulate a world in which stories take place and structure a narrative around them. We distinguish these from naive, purely agent-based world simulations from which a narrative emerges as a side-effect, such as \textit{Dwarf Fortress}. Generators in this category have some degree of intentionality in causing a story to occur. Examples include Versu's agent-based storytelling  \cite{evans2014versu}, or the AI Storyteller system in RimWorld. Generated narratives may serve as motivation for player or non-player characters to act, may reveal exposition to lead to further game events, or may be an end in itself for the player, whereby the narrative's resolution is the ultimate aim of the game (such as in a choose-your-own-adventure).

\subsubsection{Aesthetic \& Decorative Elements}
Generators which produce thematic elements, decorations, visual and aural content that augment and style a game in a particular way. This content may not specifically impact a challenge for the player (if, indeed, the game is designed around the notion of challenges or tests) but may contribute to the game's general atmosphere and mood.

\section{PCG-Based Game Design Patterns}
In \cite{treanor2015} the authors describe a collection of design patterns for taking AI techniques and using them as the basis for a game. In this section we extend these design patterns with additional, specialised patterns that target concepts in procedural content generation specifically. In some cases these are variations or blends of design patterns from the original paper.

%\subsection{AI Is Editable}
%\textbf{Original Pattern:} `\textit{Have the player directly change the elements of an AI agent that is central to gameplay.}'\\
%\textbf{Explanation:} By allowing the player to directly manipulate the parameters and settings of a procedural generator, they are effectively allowed to `edit' the generative space the system occupies, and thus indirectly edit the output of the system. This allows challenges and creative play to be designed around the higher-level generative space rather than the specific content it produces. Editing the generator might be a limited resource if the editing process is very powerful, or it might be unlimited if the task of finding the correct edits is itself a challenging task.\\
%\textbf{Example:} \textit{Endless Web} \cite{endlessweb} is an implicit example of this pattern, where the player is editing the generator's parameters through the choices they make during gameplay. \textit{CLAY}\footnote{http://cutgarnetgames.itch.io/clay} is a more explicit example of this pattern, in which the player is given several buttons to press that directly control parameters in the game. They can use these at any point during gameplay, as often as they like, and immediately visualise the result of their actions.
%
%\subsection{Player As Objective Function}
%
\subsection{AI As Creative Proxy}
\textbf{Pattern:} The player designs or tweaks a generative system which then goes on to produce content, either for mechanical or aesthetic purposes.\\
\textbf{Explanation:} Instead of directly designing a piece of content co-operatively with an AI system (as in the \textit{AI as Co-Creator} AI-based game pattern), in this pattern the player designs a system that will act as a creator of other content, and that system then goes on to have a role within the game. This might be a purely aesthetic, playful or non-critical role, or it may have a mechanical purpose. A key aim of this pattern is to get the player to engage in the meta-level creation process, designing a generator while thinking about the kind of generative space they are defining in doing so.\\
\textbf{Example:} \cite{saunders} describes a system of AI agents acting as creative communities with some agents acting as critics, some as creators, and some as gatekeepers that filter art between communities and set trends. A game in which the player designs an artist which then enters such a creative community would task the player with thinking abstractly about a space of art, rather than a single piece on its own. 

\subsection{AI As Meta-Environment}
\textbf{Pattern:} The possibility space of a generative system acts as a space the player can travel through using transformative operations.\\
\textbf{Explanation:} Akin to travelling through physical game space to solve problems, explore areas or reach objectives, in this pattern the player travels through the abstract generative space of a procedural generator, by making adjustments to the generator such as altering parameters or input data. The player's aim might be to produce a particular example of content, have the generator occupy a particular region of space, or achieve some other in-game goal that is affected by the state of the generator. The adjustments to the generator can be thought of as edges connecting vertices in a graph, which represent distinct states of the generator.\\
\textbf{Example:} In \textit{Endless Web} the player travels through physical game space when solving levels. The exit they choose to each level effectively allows them to travel in the possibility space of the level generator, by altering generator parameters that affect what kind of levels are generated subsequently. Figure \ref{ewshot} shows the `world map' showing what state the generator is currently in and where the player can move to in the graph of generator states.

\subsection{AI Is Filtered}
\textbf{Pattern:} The player acts as a fitness function or filter for generated content.\\
\textbf{Explanation:} In this pattern, a generator produces content which has some role in the game's systems. The player can control the generator through selecting, ranking or filtering its content in some way. This might be similar to an interactive evolutionary system where the player selects content which feeds back into the system, or it might be a culling process where the player takes the role of the `tester' in a `generate and test' process.\\
\textbf{Example:} Interactive evolutionary games like \textit{Petalz} generate flower designs which the player filters by selecting flowers to breed together. Another unintentional example of this pattern is `scumming', a technique developed by roguelike players where certain randomised events can be re-triggered until a favourable outcome is found. Although this is not part of the design, its emergence as a play technique is an interesting example of this behaviour.

\subsection{AI Is Interrupted}
\textbf{Pattern:} The player interjects in the execution of generative systems, stopping, slowing or restarting their progress.\\
\textbf{Explanation:} This pattern takes generative systems, possibly expressed as multi-agent systems, and allows the player to interrupt them partway through their execution in order to achieve a particular effect. This might be in order to take advantage of partially-generated content when it reaches a stage that the player deems useful, or it might be to reconfigure the environment so the process can be restarted and continues in a different way.\\
\textbf{Example:} Although no games specifically exemplify this pattern, games such as \textit{Lemmings} or \textit{Dungeon Keeper} essentially express simply generative systems through autonomous agents that perform actions to modify the game world. The player can alter these agents to change the generative processes in order to solve puzzles.

\section{Case Study: Sliding Doors}
\textit{Sliding Doors} is a choose-your-own-adventure game in which the player 
controls a character through a story, making decisions at various points to decide what the player does next, thereby influencing the chain of events that transpires. Typically, games of this type use human-authored stories where the narrative branches off depending on choices but often meet up again in the future to simplify the number of possible stories. Examples of games of this type include Telltale's \textit{The Walking Dead}. Some games use story-generation systems to produce more variety in their storylines - \textit{Versu} models story characters as agents and allows relationships between them to combine with a loose narrative structure to generate stories.

\begin{figure}
\begin{center}
\includegraphics[width=0.9\columnwidth]{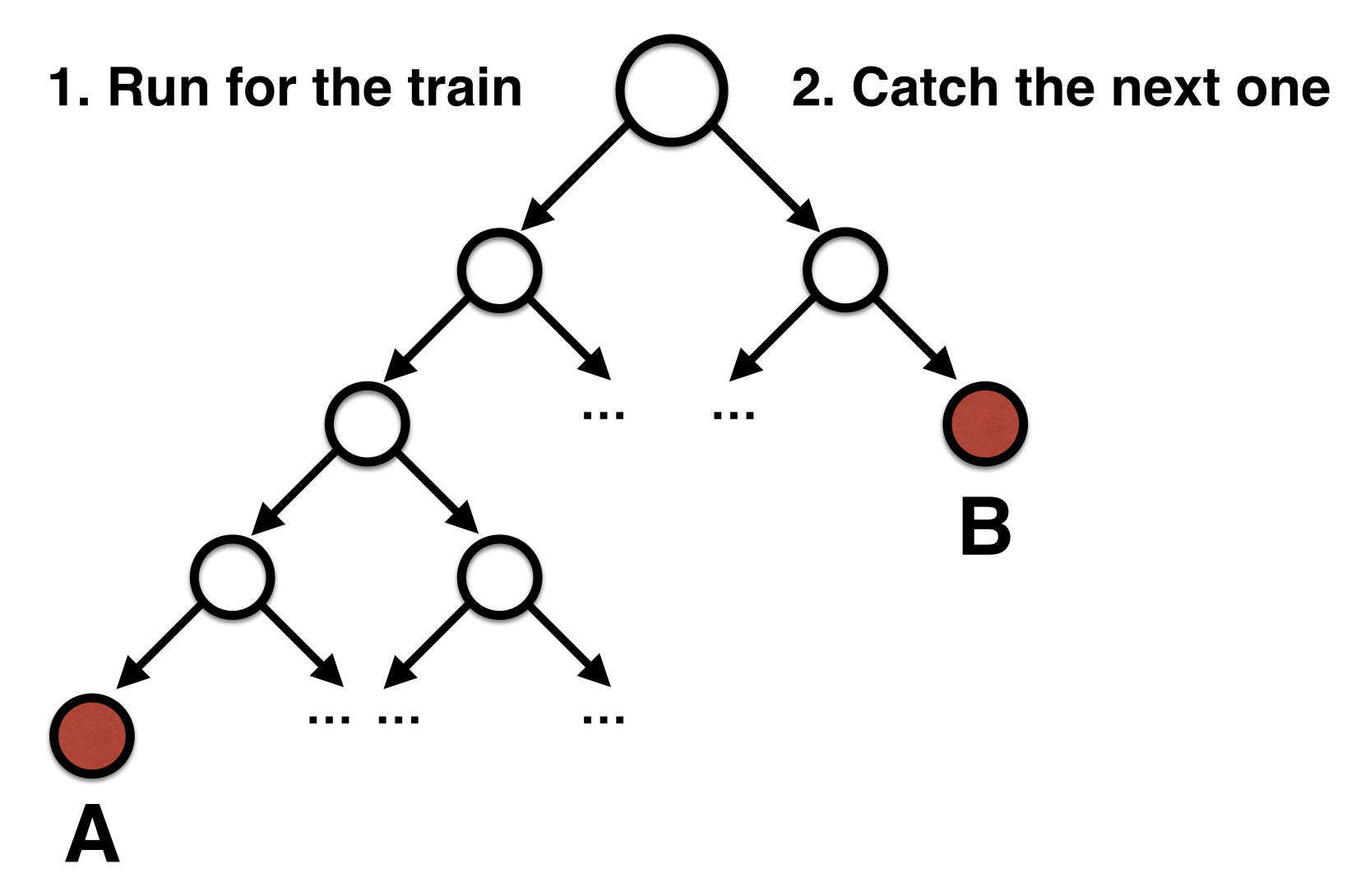}
\end{center}
\caption{An example decision point in \textit{Sliding Doors}. The left and right subtrees represent (partially represented) sub-stories resulting in taking either the first or second decision respectively. The coloured node in each subtree represent a point in the future which the player has chosen to reveal.} 
\label{choiceexample}
\end{figure}

Our concept for Sliding Doors utilises a story generator to create a narrative of events during which the player is regularly posed with scenarios and asked to make choices. In a normal game this decision would be made only using the current context, however in Sliding Doors the player can choose to view some future state in the timelines represented by each individual choice (a binary choice presents the player with two possible future visions, for example, one for each choice). The player can choose how far into the future each vision comes from, however the nearer to the present time they choose to look at, the more vague the vision they receive.

As an example, consider the following scenario based on the movie \textit{Sliding Doors}, which this game is named after. The main character reaches a train station to discover she is about to miss her train. The player is offered a choice: try and run for the train, or miss it and be late for work. In our story generator, this is a branching point in a larger tree representing the entire space of the current story. Each choice represents a subtree from the story node the player is currently in, as shown in Figure \ref{choiceexample}, where the left and right subtrees represent the game progressing after the player chooses either the first or second choice.

In Figure \ref{choiceexample} the player has chosen one point in each subtree to reveal information about. In the first subtree the point is quite far into the future - this will increase the detail in the vision. The player might have specific information revealed to them, such as a view of them in a particular location, a time or date, even some lines of dialogue. The value of the additional detail is offset by the fact that, at a depth of four nodes into the future, it is one of eight possible futures that the story could reach (assuming binary choices - if the choices are more complex, the number of possible futures is even higher). This means that statistically it is less likely that this will actually happen, which changes the value of the revealed information. By contrast, in the second subtree the player chooses a more near-future node to reveal. Because this node is more reliable (being only two nodes deep in the tree) the information revealed is far more vague. Perhaps the player sees their character in tears, but is unable to tell where they are or why they are crying. 

In the language of our earlier taxonomies, this design sketch describes a game which allows the player to view (and prune, through their choices) the generator's \textit{State Space} in order to make decisions affecting the game's \textit{Narrative}. The structure of the generator's state space is partially represented in-game by allowing the player to select the depth into the future they wish to examine, although the specifics of branching story nodes remain hidden.

One of the appeals of story-driven games such as this is that the player does not know the future when making decisions, and thus uncovers a narrative as they make choices throughout the game (and may never know the consequences of choices they didn't make). Because we are explicitly breaking this convention by allowing the player to view the future states of the story, some kind of uncertainty and limitation is introduced to make the player's decisions less straightforward. Viewing points further in the future naturally is less reliable because of the branching factor of the story, but points in the near future are still reliable, so the introduction of more vague results from story nodes closer to the present time is intended to provide a counterweight to the strength of knowing something about a story event that is happening soon.

We plan to implement a prototype version of Sliding Doors in the near future, possibly as an elaboration of an existing story generation system. It is worth noting that \textit{Until Dawn}, a AAA PlayStation 4 game, employs a hint system in its static, human-authored narrative that has a similar way of referencing the future. Players can find objects in certain story branches that indicate the branch they are on contains certain possible futures (such as the death of a character). By building this idea into a generative system, and adding a notion of player choice and information tradeoffs, Sliding Doors shows how such an idea can be built into the very core of a game design, foregrounding a PCG-based game mechanic.

\begin{figure*}
\includegraphics[height=0.7\columnwidth]{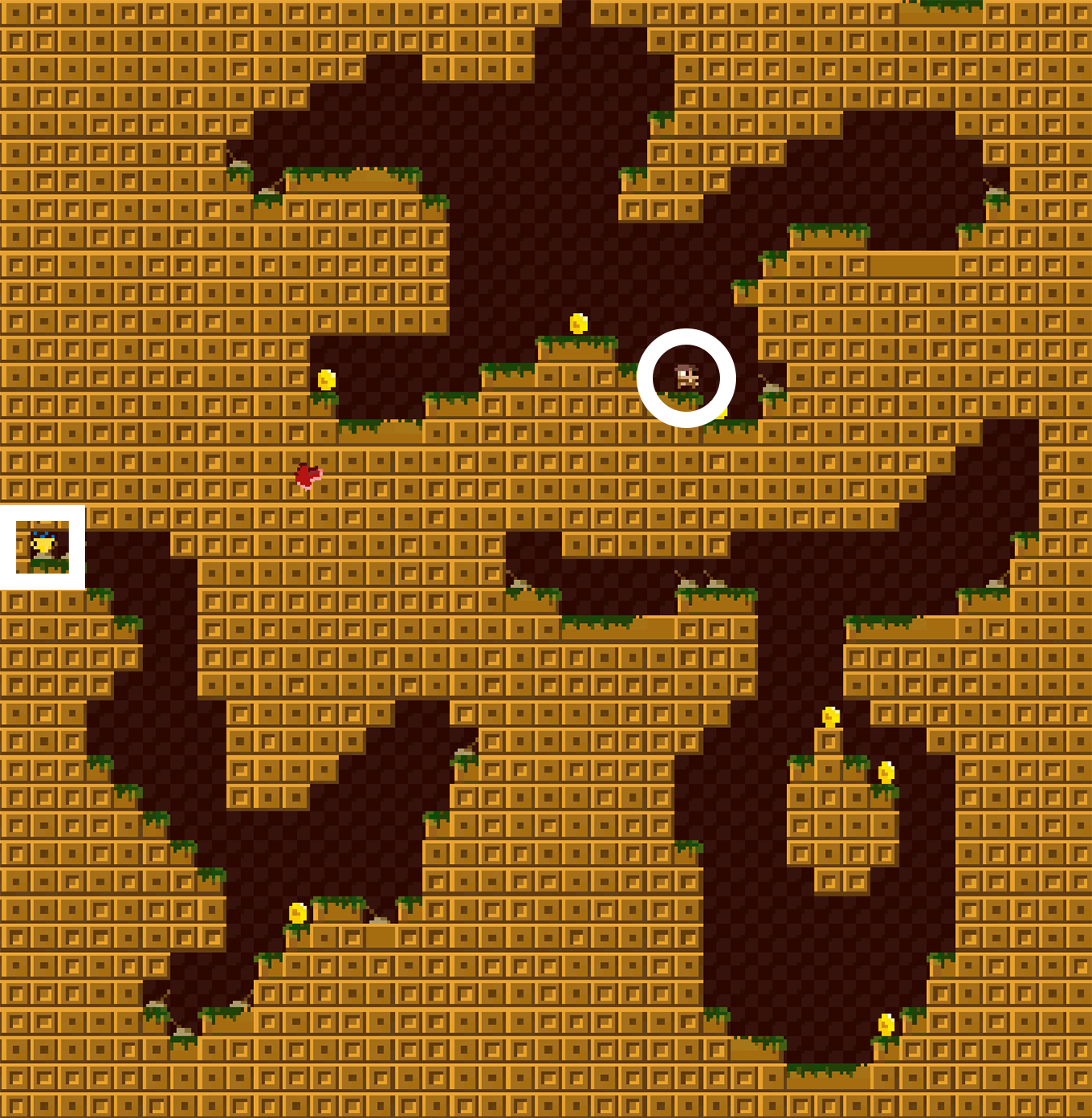}\hfill
\includegraphics[height=0.7\columnwidth]{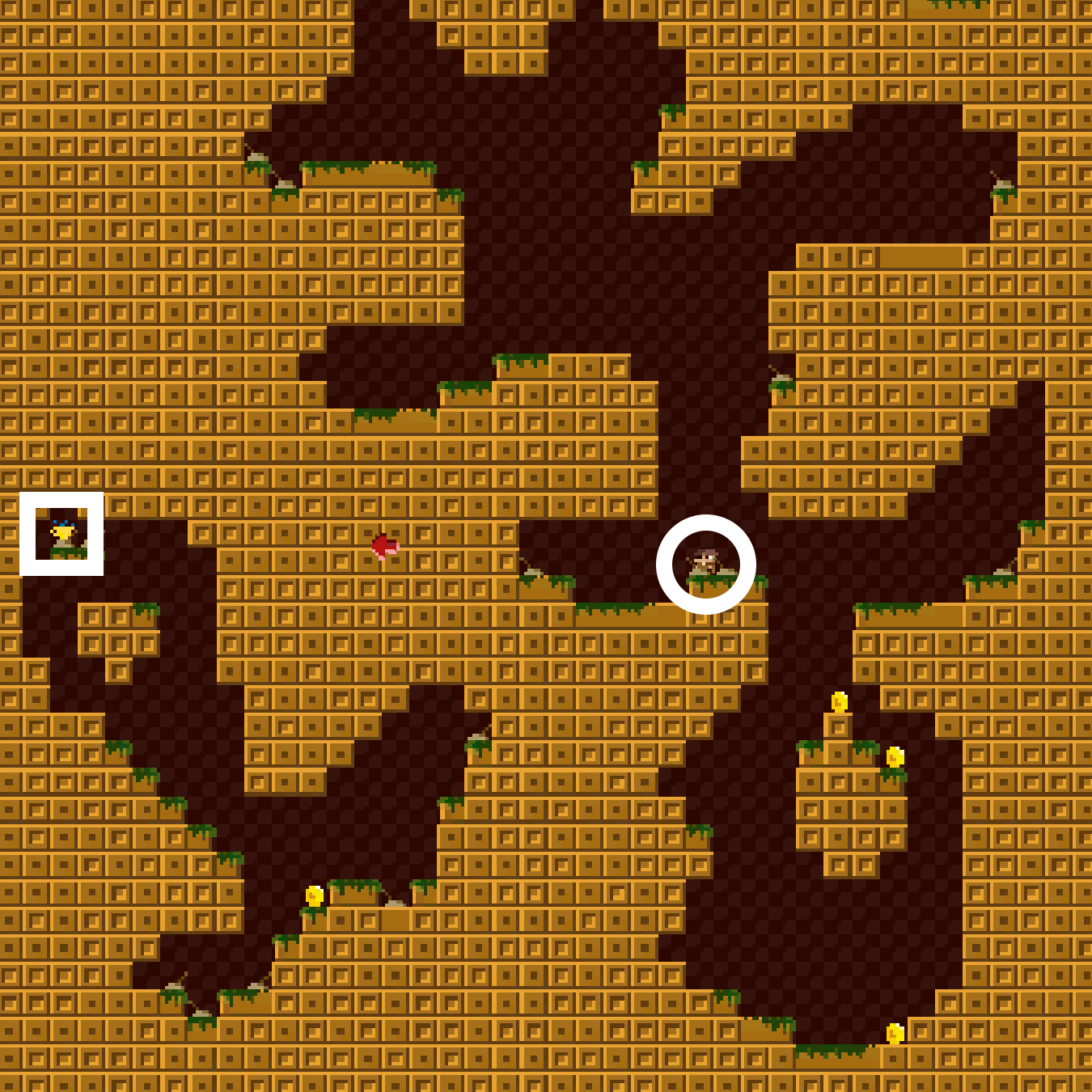}\hfill
\includegraphics[height=0.7\columnwidth]{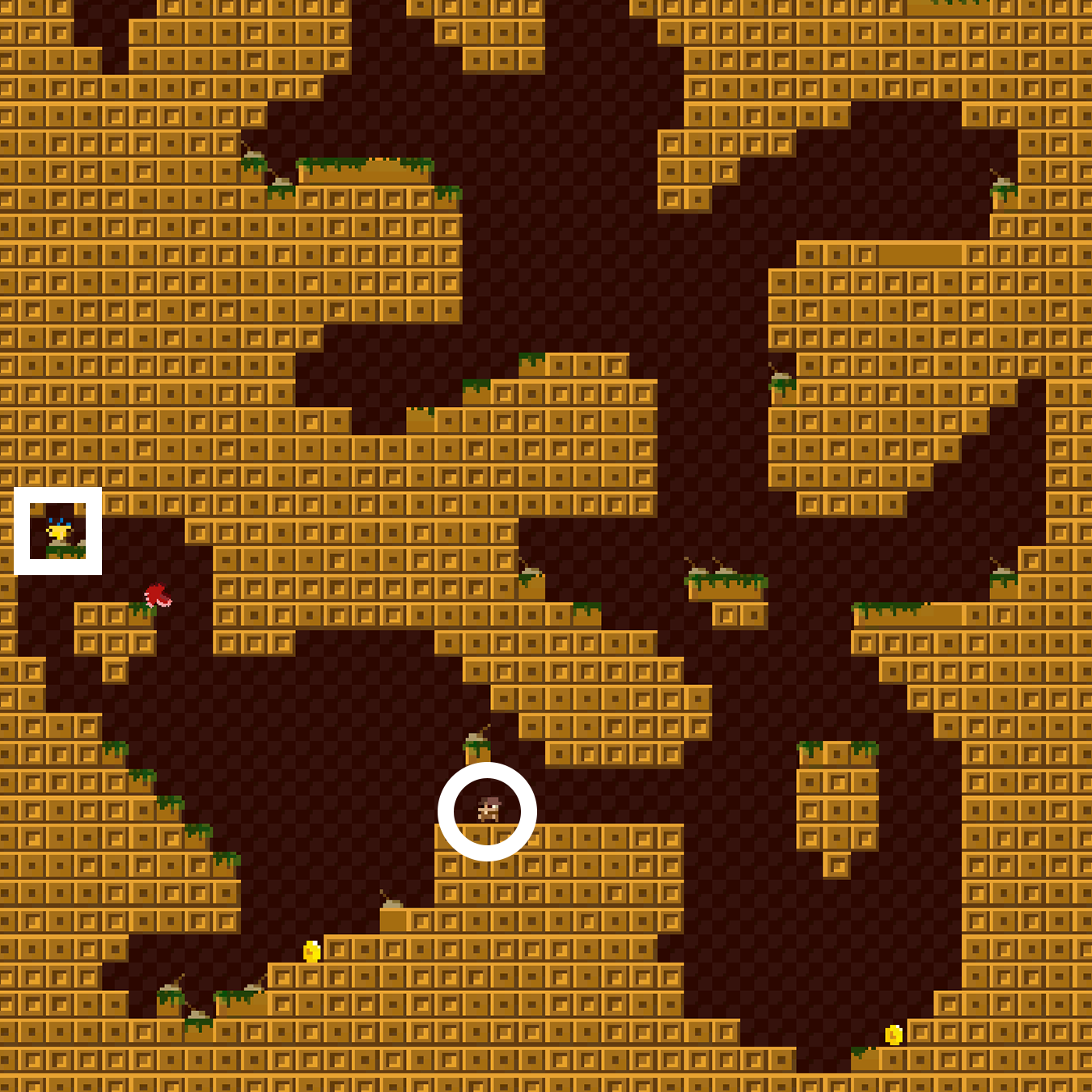}
\caption{Three stages of a level in \textit{Tombs Of Tomeria} being solved, zoomed out to show the level. The circle marks the player position, and the square marks the goal. At each stage the player has adjusted a generator parameter, opening up more space in the generated level.}
\label{tombschange}
\end{figure*}

\section{Case Study: Tombs Of Tomeria}
Tombs of Tomeria \cite{tombsclay} is a prototype adventure game in which the player can reshape levels in order to access new areas, discover secrets, and make progress. The player achieves this by reaching levers scattered throughout levels and switching them back and forth. Each lever is a binary state -- switched either left or right -- and flipping the lever causes some blocks in the level to disappear or new ones to reappear. In the current version of the game the player must seek an exit to each level by flipping levers back and forth, while collecting as many artefacts as possible along the way.

Tombs' levels are generated using a cellular automata. Each level begins with a random seed and a randomly initialised grid of blocks, and is then processed according to rules similar to Conway's Game Of Life. Parameters govern crucial features of the level generation process -- in particular, the initial random distribution of blocks (Initial Random Chance, or IRC), and the number of times the level is iterated upon (Number Of Iterations, or NOI -- more iterations produces a smoother and sparser level). 

Whenever the player flips a lever in Tombs, one of the two previously-mentioned parameters is adjusted by some delta set by the designer (in this case, the random change is modified by 0.5\% or the number of iterations is changed by 1). The level is then regenerated using the same initial random seed, meaning that the state of the level before and after the lever switch is close enough that one appears to be a slight variation of the other. This gives the impression of the level environment slowly `shifting' as levers are pulled. Figure \ref{tombschange} shows a simple level being solved in three stages.

%\begin{figure}
%\begin{center}
%\begin{subfigure}[b]{\columnwidth}
%\centering
%\includegraphics[width=0.7\columnwidth]{images/tombs1.png}
%\end{subfigure}
%
%\begin{subfigure}[b]{\columnwidth}
%\centering
%\includegraphics[width=0.7\columnwidth]{images/tombs2.png}
%\end{subfigure}
%
%\begin{subfigure}[b]{\columnwidth}
%\centering
%\includegraphics[width=0.7\columnwidth]{images/tombs3.png}
%\end{subfigure}
%\end{center}
%\caption{Three stages of a level in \textit{Tombs Of Tomeria} being solved, zoomed out to show the level. The circle marks the player position, and the square marks the goal. At each stage the player has adjusted a generator parameter, opening up more space in the generated level.}
%\label{tombschange}
%\end{figure}

Because the size of the change caused by a single lever is known in advance, and the minimum and maximum changes can be calculated based on how many levers in total are in the level, Tombs can design levels and check their solvability in advance. It does this by generating a level for every possible set of lever states (assuming a constant random seed, and a known initial set of parameter values). Then, given an initial location for the player, Tombs can simulate gameplay by finding all accessible levers and recursively flipping levers and exploring the resulting levels to create a graph of mutually accessible game states. This lets Tombs calculate how much of the level space can be explored by the player, and where treasure and level exits can be placed to be challenging but solvable.

In the language of our earlier taxonomies, Tombs allows the player access to the generator's \textit{Parameters} in order to affect the generation of the \textit{Environment}. It does this explicitly, to some degree - the player is aware they are interacting with a system by switching levers, but may not understand exactly what the impact of their actions are (beyond a vague sense that the level is getting more dense or more sparse when a lever is thrown). 

The use of levers in the design is intended to simplify and clarify interactions with the generator for the player. Because only one lever can be thrown with a single action, the possible changes made by the player are restricted to a single parameter at at time. This has several useful effects on the game design: firstly, the scope of a single player action is minimised which makes it easier for the player to relate pulling a lever with its effect on the level; secondly, it simplifies the space of possible player actions which simplifies the automated level analysis process described above; finally, it helps limit the total power the player has over the parameter space, because the number of levers defines the maximum a parameter can vary from the starting value.

\section{Discussion}
Throughout the case studies presented in this paper, as well as our surveys of existing work and games in this area, several commonly recurring themes emerged that we believe are important to future research in this area as well as for the future development of PCG-based games. We outline them below.

\subsection{Visualisation} 
Accurately imagining the space of a generator's output, or predicting the effect of changing something about a generator, is exceptionally hard even for an expert practitioner of generative systems. Thus, games which ask the player to make strategic, deductive or creative decisions on the basis of an interaction with a generator need to be able to easily visualise the results of such interactions. Sometimes this visualisation is relatively straightforward -- in the case of \textit{Sliding Doors}, for example, the player is shown a partial scene from a future branch of the story space when they select it. In other games this may be less simple to do. In an early prototype of \textit{Tombs} the player was not able to know the effect a lever had without pulling it, which made it hard for the player to make decisions about level traversal.

In the case of viewing the effects of a single change, overlaying or previewing the effects of a change may be enough to let the player decide what to do next (in \textit{Tombs}, blocks which will change when a lever is pulled might discolour when the player walks over it, providing a diegetic way of seeing two outputs from the generator simultaneously. PCG-based games that require the player to make judgements about entire spaces of content, or multiple generators, may benefit from ideas from analytical PCG research, such as Expressive Range Analysis \cite{era}.
 
\subsection{Managing Control}
As we noted in our taxonomy section, procedural generators are typically tasked with important roles in a game's systems, such as controlling player progression or defining the environment the player moves through and explores. Procedural generators are frequently tasked with designing part of the game experience, and therefore giving the player control over a generator potentially empowers them to change the game outside of the designer's expectations or desires.

This effect may be desirable in some cases, but for the most part we anticipate that an important part of designing PCG-based games will be knowing how and when to limit player control over a system. In \textit{Sliding Doors}, interaction with the story generator is a limited resource (the player can investigate one node per branch per decision), and the knowledge gained from such interactions is imperfect. In \textit{Tombs}, the player can only interact with the generator at specified places in the game environment, and the same small degree of change is made in every interaction. While some of these approaches may be overly conservative, they show how a balance can be struck between the wild freedom of a generator and the intended direction of a game's design.

\subsection{Encountering Generative Concepts}
PCG-based games represent an exciting area in which to do game design, to develop new technology, and to explore new ideas about interactivity with generators. However, we also believe they represent an opportunity for people to gain more familiarity with generative software in a playful context, without needing to learn how to code. Generators benefit from being taught in an interactive context, and building this context into a game is an interesting way of teaching and explaining some of these ideas. 

%Concepts such as possibility space, discoverability, uniqueness and novelty are often discussed but poorly understood in discourse about procedural generation in games. 
%
% This is not only a problem for developers, critics and academics, but it also stifles experimentation and creativity in the applications of generative techniques to games. PCG is often seen as simply a source of random, meaningless fodder content to pad out games \cite{rogers}, but this is a self-fulfilling prophecy -- this perception puts off those who might experiment further, and  
 
Procedural generation is often seen as little more than a cheap source of randomness to pad a game out with additional content, rather than an artistic tool capable of helping a designer achieve specific goals in a work. Stereotypes about procedural generation are constantly self-reinforcing, with programming skills often being an additional barrier to entry (although not universally\footnote{http://www.tracery.io}). The discipline would benefit from more ways to motivate, interest and excite people about generative systems, so that the area can grow and develop into an accessible and diverse medium. We hope that PCG-based games can contribute to this in a small way.

%While PCG-based games represent an exciting area in which do game design and to develop technology, we believe they also represent an opportunity for people to gain familiarity with concepts relating to generative systems in a playful concept. These games all focus on some aspect of a generator and allow the player to adjust and explore it, whether that is its possibility space, the consequences of rules, the effects of parameters. Games designed around these ideas can play with notions of discoverability, uniqueness, adjacency of points in generative space, and more.
%
%We believe that such games would not only be powerful ways for players to encounter these concepts and gain familiarity with them, but also ways to help people appreciate generative systems more. Many people already enjoy the output of generative software, such as the thousands of bots that populate social media, many of whom have tens of thousands of followers. Games offer another way for people to engage with and appreciate the so-called `generative aesthetic'.

\section{Conclusions}
In this paper we refined the notion of a PCG-based game, a direct extension of AI-based games where a procedural generator is a central part of the player's experience and interactions with the game. We discussed the history of the concept, including AI-based games and prototypical work in the area, and general theory relating to procedural generators and people's interaction with them. We then discussed ways of taxonomising procedural generators in terms of their affordances for game designers, and linked this to a series of design patterns inspired by AI-based design patterns. Finally, we gave details of several case studies, including implemented prototypes, of PCG-based games that illustrate interesting features of the concept.

There are many examples of games that skirt this exciting area of game design, and we hope to see more prototypes and projects working in this domain in the future. By finding new uses and purposes for generative systems in games, we can move away from stereotypical applications and find exciting and diverse new uses instead. Additionally, we hope to see more refinement of subcategories of AI-based games -- the case studies in this paper all owe a debt to the design patterns and ideation described in \cite{treanor2015}, and we hope this paper demonstrates the usefulness and potential of the approach described. We believe it can be applied with similarly positive results on other AI techniques.
%
%\section{Acknowledgements}
%This work would not have been possible without the BIRS Seminar \textit{Computational Modelling in Games}. This work was also supported in part by EPSRC Grant XX/XXX, and the Virtual Institute for Computational Expression.

\bibliographystyle{aaai}
\bibliography{citations-mpe-eds-2016-06-06}

\end{document}